\documentclass[a4,a4wide]{article}

\usepackage{aaai2}
\usepackage{times}
\usepackage{helvet}
\usepackage{courier}

\usepackage{amssymb}
\usepackage{latexsym}

\usepackage{subfig}

\usepackage{tikz}
\usetikzlibrary{backgrounds}
\usetikzlibrary{positioning}
\usetikzlibrary{automata}
\usetikzlibrary{trees}

\newtheorem{example}{Example}
\newtheorem{definition}{Definition}
\newtheorem{proposition}{Proposition}

\newcommand{\nat}{{\mathbb N}}

\newcommand{\mysat}{\mathrel|\joinrel\sim}
\newcommand{\mynonsat}{\mathrel|\joinrel\not\sim}

\begin{document}

\title{Analysis of Dialogical Argumentation\\ via Finite State Machines}

\author{Anthony Hunter\\
	Department of Computer Science,\\
		University College London, \\
		Gower Street, London WC1E 6BT, UK}

\maketitle

\begin{abstract}
Dialogical argumentation is an important cognitive activity by which agents exchange arguments and counterarguments as part of some process such as discussion, debate, persuasion and negotiation. Whilst numerous formal systems have been proposed, there is a lack of frameworks for implementing and evaluating these proposals. First-order executable logic has been proposed as a general framework for specifying and analysing dialogical argumentation. In this paper\footnote{This paper has already been published in the Proceedings of the International Conference on Scalable Uncertainty Management (SUM'13), LNCS 8078, Pages 1-14, Springer, 2013.}, we investigate how we can implement systems for dialogical argumentation using propositional executable logic. Our approach is to present and evaluate an algorithm that generates a finite state machine that reflects a propositional executable logic specification for a dialogical argumentation together with an initial state. We also consider how the finite state machines can be analysed, with the minimax strategy being used as an illustration of the kinds of empirical analysis that can be undertaken.
\end{abstract}

\section{Introduction}\label{section:introduction}

Dialogical argumentation involves 
agents exchanging arguments in activities such as discussion, debate, persuasion, and negotiation \cite{BHbook}. 
Dialogue games are now a common approach to characterizing argumentation-based agent dialogues 
(e.g. \cite{AMP00,BH09,dignum:00,FT11,Ham71,Mac79,MP02b,MEPA03,parsons:03a,prakken:05,walton:95}). 
Dialogue games are normally made up of a set of communicative acts called moves, and a protocol specifying which moves can be made at each step of the dialogue. 
In order to compare and evaluate dialogical argumentation systems, we proposed in a previous paper that first-order executable logic could be used as common theoretical framework to specify and analyse dialogical argumentation systems \cite{BH12}.

In this paper, we explore the implementation of dialogical argumentation systems in executable logic. 
For this, we focus on propositional executable logic as a special case, 
and investigate how a finite state machine (FSM) can be generated as a representation of 
the possible dialogues that can emanate from an initial state.
The FSM is a useful structure for investigating various properties of the dialogue, including conformance to protocols, and application of strategies.
We provide empirical results on generating FSMs for dialogical argumentation, and how they can be analysed using the minimax strategy. 
We demonstrate through preliminary implementation that it is computationally viable to generate the FSMs and to analyse them. This has wider implications in using executable logic for applying dialogical argumentation in practical uncertainty management applications, since we can now empirically investigate the performance of the systems in handling inconsistency in data and knowledge.

\section{Propositional executable logic}\label{section:propositional}

In this section, we present a propositional version of the executable logic which we will show is amenable to implementation. This is a simplified version of the framework for first-order executable logic in \cite{BH12}.

We assume a set of atoms which we use to form propositional formulae in the usual way using disjunction, conjunction, and negation connectives.
We construct modal formulae using the $\boxplus$, $\boxminus$, $\oplus$, and $\ominus$ modal operators.
We only allow literals to be in the scope of a modal operator.
If $\alpha$ is a literal, then each of $\oplus\alpha$, $\ominus\alpha$, $\boxplus\alpha$, and $\boxminus\alpha$ is an {\bf action unit}.
Informally, we describe the meaning of action units as follows: 
$\oplus\alpha$ means that the action by an agent is to add the literal $\alpha$ to its next private state; 
$\ominus\alpha$ means that the action by an agent is to delete the literal  $\alpha$ from its next private state;  
$\boxplus\alpha$ means that the action by an agent is to add  the literal $\alpha$ to the next public state; 
and $\boxminus\alpha$ means that the action by an agent is to delete  the literal $\alpha$ from the next public state.

We use the action units to form {\bf action formulae} 
as follows using the disjunction and conjunction connectives:
(1) If $\phi$ is an action unit, then $\phi$ is an action formula;
And 
(2) If $\alpha$ and $\beta$ are action formulae, then $\alpha\vee\beta$ and $\alpha \wedge\beta$ are action formulae.
Then, we define the action rules as follows: 
If $\phi$ is a classical formula and $\psi$ is an action formula 
then $\phi\Rightarrow\psi$ is an {\bf action rule}.
For instance,  $\tt b(a) \Rightarrow \boxplus c(a)$ is an action rule (which we might use in an example where $\tt b$ denotes belief, and $\tt c$ denotes claim, and $\tt a$ is some information). 

Implicit in the definitions for the language is the fact that we can use it as a meta-language \cite{WMP05}. 
For this, the object-language will be represented by terms in this meta-language. 
For instance, the object-level formula $\tt p(a,b) \rightarrow q(a,b)$ can be represented by a term where the object-level literals $\tt p(a,b)$ and $\tt q(a,b)$ are represented by constant symbols, and $\rightarrow$ is represented by a function symbol. 
Then we can form the atom $\tt belief(p(a,b) \rightarrow q(a,b))$ where $\tt belief$ is a predicate symbol.
Note, in general, no special meaning is ascribed the predicate symbols or terms.
They are used as in classical logic. 
Also, the terms and predicates are all ground, and so it is essentially a propositional language.

We use a state-based model of dialogical argumentation with the following definition of an execution state.
To simplify the presentation, we restrict consideration in this paper to two agents. 
An execution represents a finite or infinite sequence of execution states. 
If the sequence is finite, then $t$ denotes the terminal state, otherwise $t = \infty$.

\begin{definition}
An {\bf execution}  $e$ is a tuple $e = (s_1,a_1,p,a_2,s_2,t)$, 
where for each $n \in \nat$ where $0 \leq n \leq t$, 
$s_1(n)$ is a set of ground literals,
$a_1(n)$ is a set of ground action units,
$p(n)$ is a set of ground literals,
$a_2(n)$ is a set of ground action units,
$s_2(n)$ is a set of ground literals,
and $t \in \nat \cup \{ \infty \}$.
For each $n \in \nat$, if $0 \leq n \leq t$, then an {\bf execution state} is $e(n)$ = $(s_1(n),a_1(n),p(n),a_2(n),s_2(n))$
where $e(0)$ is the {\bf initial state}. 
We assume $a_1(0)$ = $a_2(0)$ = $\emptyset$.
We call 
$s_1(n)$ the private state of agent 1 at time $n$,
$a_1(n)$ the action state of agent 1 at time $n$,
$p(n)$ the public state at time $n$,
$a_2(n)$ the action state of agent 2 at time $n$,
$s_2(n)$ the private state of agent 2 at time $n$.
\end{definition}

In general, there is no restriction on the literals that can appear in the private and public state. The choice depends on the specific dialogical argumentation we want to specify. This flexibility means we can capture diverse kinds of information in the private state about agents by assuming predicate symbols for their own beliefs, objectives, preferences, arguments, etc, and for what they know about other agents. The flexibility also means we can capture diverse information in the public state about moves made, commitments made, etc.

\begin{example}
\label{ex:turn:2}
The first 5 steps of an infinite execution where each row in the table is an execution state 
where $\tt b$ denotes belief, and $\tt c$ denotes claim. 
\begin{center}
\begin{tabular}{|l|l|l|l|l|l|}
\hline
$n$ & $s_1(n)$  & $a_1(n)$ & $p(n)$ & $a_2(n)$  & $s_2(n)$  \\
\hline
\hline
0 &  $\tt b(a)$	&  & & & $\tt b(\neg a) $ \\
\hline
1 &  $\tt b(a)$	& $\tt \boxplus c(a)$ &  & & $\tt b(\neg a) $ \\
&& $\tt \boxminus c(\neg a)$ &&& \\
\hline
2 &  $\tt b(a)$	& & $\tt c(a)$ & $\tt \boxplus c(\neg a)$  & $\tt b(\neg a) $  \\
  &   & &   &  $\tt \boxminus c(a)$ & $\tt b(\neg a) $  \\
\hline
3 &  $\tt b(a)$	& $\tt \boxplus c(a)$  & $\tt c(\neg a)$ & & $\tt b(\neg a) $ \\
 &   	&  $\tt \boxminus c(\neg a)$ &   & &   \\
\hline
4 &  $\tt b(a)$	& & $\tt c(a)$ &  $\tt \boxplus c(\neg a)$   & $\tt b(\neg a) $  \\
  &   	& &   &   $\tt \boxminus c(a)$  &    \\
\hline
5 & $\dots$  & $\dots$ & $\dots$ & $\dots$  & $\dots$  \\
\hline
\end{tabular}
\end{center}
\end{example}

We define a system in terms of the action rules for each agent, which specify what moves the agent can potentially make based on the current state of the dialogue. In this paper, we assume agents take turns, and at each time point the actions are from the head of just one rule (as defined in the rest of this section).

\begin{definition}
A {\bf system} is a tuple $(Rules_x, Initials)$
where  
$Rules_x$ is the set of action rules for agent $x \in \{ 1, 2 \}$,
and $Initials$ is the set of initial states.
\end{definition}

Given the current state of an execution, the following definition captures which rules are fired. 
For agent $x$, these are the rules that have the condition literals satisfied by the current private state $s_x(n)$ and public state $p(n)$. 
We use classical entailment, denoted $\models$, for satisfaction, but other relations could be used (e.g. Belnap's four valued logic).
In order to relate an action state in an execution with an action formula, we require the following definition.

\begin{definition}
For an action state $a_x(n)$, and an action formula $\phi$, $a_x(n)$ {\bf satisfies} $\phi$,  denoted $a_x(n) \mysat \phi$, as follows.
\begin{enumerate}
\item $a_x(n) \mysat \alpha$ iff $\alpha \in a_x(n)$ when $\alpha$ is an action unit
\item $a_x(n) \mysat  \alpha \wedge \beta$ iff $a_x(n) \mysat \alpha$ and $a_x(n) \mysat \beta$
\item $a_x(n) \mysat  \alpha \vee \beta$ iff $a_x(n) \mysat \alpha$ or $a_x(n) \mysat \beta$
\end{enumerate}
For an action state $a_x(n)$, and an action formula $\phi$,
$a_x(n)$ {\bf minimally satisfies} $\phi$, 
denoted $a_x(n) \Vdash \phi$,
iff $a_x(n) \mysat \phi$ 
and for all $X \subset a_x(n)$, $X \mynonsat \phi$.
\end{definition}

\begin{example}
Consider the execution in Example \ref{ex:turn:2}. For agent 1 at n = 1, 
we have $a_1(1) \Vdash {\tt \boxplus c(a) \wedge \boxminus c(\neg a)}$.
\end{example}

We give two constraints on an execution to ensure that they are well-behaved. The first (propagated) ensures that each subsequent private state (respectively each subsequent public state) is the current private state (respectively current public state) for the agent updated by the actions given in the action state. The second (engaged) ensures that an execution does not have one state with no actions followed immediately by another state with no actions (otherwise the dialogue can lapse) except at the end of the dialogue where neither agent has further actions.

\begin{definition}
\label{def:execution}
An execution $(s_1,a_1,p,a_2,s_2,t)$ is {\bf propagated} 
iff for all $x \in \{1,2\}$, for all $n \in \{0, \ldots, t-1\}$, where $a(n) = a_1(n) \cup a_2(n)$
\begin{enumerate}

\item $s_x(n+1) = ( s_x(n) \setminus \{\phi \mid \ominus\phi \in a_x(n) \} ) \cup \{ \phi \mid \oplus \phi \in a_x(n) \}$

\item $p(n+1) = ( p(n) \setminus \{\phi \mid \boxminus\phi \in a(n) \}) \cup \{ \phi \mid \boxplus \phi \in a(n) \}$

\end{enumerate}
\end{definition}

\begin{definition}
\label{def:execution}
Let $e = (s_1,a_1,p,a_2,s_2,t)$ be an execution and $a(n) = a_1(n) \cup a_2(n)$. 
$e$ is {\bf finitely engaged} 
iff (1) $t \neq \infty$; 
(2) for all $n \in \{1, \ldots, t-2\}$,  if $a(n) = \emptyset$, then $a(n+1) \neq \emptyset$
(3) $a(t-1) = \emptyset$; 
and (4) $a(t) = \emptyset$.
$e$ is {\bf infinitely engaged} 
iff (1) $t = \infty$; 
and 
(2) for all $n \in \nat$,  if $a(n) = \emptyset$, then $a(n+1) \neq \emptyset$.
\end{definition}

The next definition shows how a system provides the initial state of an execution and the actions that can appear in an execution. It also ensures turn taking by the two agents.

\begin{definition}
\label{def:delineate}
Let $S$ = $(Rules_x, Initials)$ be a system and $e$ = $(s_1,a_1,p,a_2,s_2,t)$ be an execution.
$S$ {\bf generates} $e$ 
iff 
(1) $e$ is propogated; 
(2) $e$ is finitely engaged or infinitely engaged; 
(3) $e(0) \in Initials$; 
and (4) for all $m \in \{1, \ldots, t-1\}$
\begin{enumerate}

\item If $m$ is odd, then $a_2(m) = \emptyset$ 
and either $a_1(m) = \emptyset$ or  there is an $\phi\Rightarrow\psi \in Rules_1$ s.t. $s_1(m)\cup p(m) \models \phi$ and $a_1(m) \Vdash \psi$ 

\item If $m$ is even, then $a_1(m) = \emptyset$ 
and either $a_2(m) = \emptyset$ or  there is an $\phi\Rightarrow\psi \in Rules_2$ s.t. $s_1(m)\cup p(m) \models \phi$ and $a_2(m) \Vdash \psi$ 
\end{enumerate}
\end{definition}

\begin{example}
We can obtain the execution in Example \ref{ex:turn:2} with the following rules: 
(1) $\tt b(a) \Rightarrow \boxplus c(a) \wedge \boxminus c(\neg a)$;
And (2) $\tt b(\neg a) \Rightarrow \boxplus c(\neg a) \wedge \boxminus c(a)$. 
\end{example}

\section{Generation of finite state machines}\label{section:generation}

In \cite{BH12}, we showed that for any executable logic system with a finite set of ground action rules, and an initial state, there is an FSM that consumes exactly the finite execution sequences of the system for that initial state. 
That result assumes that each agent makes all its possible actions at each step of the execution.
Also that result only showed that there exist these FSMs, and did not give any way of obtaining them.

In this paper, we focus on propositional executable logic where the agents take it in turn, and only one head of one action rule is used, and show how we can construct an FSM that represents the set of executions for an initial state for a system. For this,  each state is a tuple $(r, s_1(n),p(n),s_2(n))$, and each letter in the alphabet is a tuple $(a_1(n),a_2(n))$, where $n$ is an execution step and $r$ is the agent holding the turn when $n < t$ and $r$ is $0$ when $n = t$.

\begin{definition}
A finite state machine (FSM) $M = (States,Trans,Start,Term,Alphabet)$ {\bf represents} a system $S$ = $(Rules_x, Initials)$ for an initial state $I \in Initials$ 
iff
\[
\begin{array}{l}
(1) States = \{ (y,s_1(n),p(n),s_2(n)) \mid \\
\hspace{2.1cm} \mbox{there is an execution $e  = (s_1,a_1,p,a_2,s_2,t)$ } \\
\hspace{2.3cm} \mbox{s.t. $S$ generates $e$ }\\
\hspace{2.3cm} \mbox{and } I = (s_1(0),a_1(0),p(0),a_2(0),s_2(0))\\
\hspace{2.1cm} \mbox{and there is an $n \leq t$ } \\
\hspace{2.3cm} \mbox{s.t. } y = 0 \mbox{ when } n = t\\
\hspace{2.3cm} \mbox{and }  y = 1 \mbox{ when } n < t \mbox{ and } n \mbox{ is odd }\\
\hspace{2.3cm} \mbox{and }  y = 2 \mbox{ when } n < t \mbox{ and } n \mbox{ is even } \}\\
\\
(2) Term = \{ (y,s_1(n),p(n),s_2(n)) \in States \mid y = 0 \}\\		
\\
(3) Alphabet = \{ (a_1(n),a_2(n)) \mid \mbox{there is an $n \leq t$ }\\ 
\hspace{2cm} \mbox{and there is an execution $e$} \\
\hspace{2.5cm} \mbox{s.t. $S$ generates $e$}\\
\hspace{2.5cm} \mbox{and } e(0) = I\\
\hspace{2.5cm} \mbox{and } e  = (s_1,a_1,p,a_2,s_2,t) \}\\
\\
(4) Start = (1,s_1(0),p(0),s_2(0))\\
\hspace{2cm} \mbox{ where } I = (s_1(0),a_1(0),p(0),a_2(0),s_2(0))
\end{array}
\] 
\mbox{\rm (5)}Trans is the smallest subset of $States\times Alphabet \times States$ s.t. for all executions $e$ and for all $n < t$ 
there is a transition $(\sigma_1,\tau,\sigma_2) \in Trans$ such that
\[
\begin{array}{l}
\sigma_1 = (x,s_1(n),p(n),s_2(n))\\
\tau = (a_1(n),a_2(n))\\
\sigma_2 = (y,s_1(n+1),p(n+1),s_2(n+1))\\
\end{array}
\]
where 
$x$ is 1 when $n$ is odd, 
$x$ is 2 when $n$ is even, 
$y$ is 1 when $n+1 < t$ and $n$ is odd, 
$y$ is 2 when $n+1 < t$ and $n$ is even, 
and $y$ is 0 when $n+1 = t$.
\end{definition}

\begin{example}
\label{ex:fsmcycle}
Let M be the following FSM where 
$\sigma_1$ = $\tt (1,\{b(a)\}, \{ \}, \{ b(\neg a) \} )$; 
$\sigma_2$ = $\tt (2, \{b(a)\},\{ c(a) \}, \{ b(\neg a) \} )$;
$\sigma_3$ = $\tt (1, \{b(a)\},\{ c(\neg a) \}, \{ b(\neg a) \} )$. 
$\tau_1$ = $(\tt \{ \boxplus c(a)$, $\tt \boxminus c(\neg a) \}, \emptyset)$; and
$\tau_2$ = $(\tt \emptyset, \{\tt \boxplus c(\neg a), \boxminus c(a) \})$.
M represents the system in Ex \ref{ex:turn:2}.
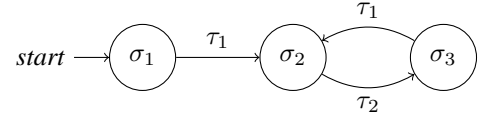
\begin{center}
\begin{tikzpicture}
\node[state,initial] (s1) at (1,1) {$\sigma_1$};
\node[state] (s2) at (3,1) {$\sigma_2$};
\node[state] (s3) at (5,1) {$\sigma_3$};
\path[->] (s1) edge node[above] {$\tau_1$} (s2);
\path[->] (s2) edge[bend right] node[below] {$\tau_2$} (s3);
\path[->] (s3) edge[bend right] node[above] {$\tau_1$} (s2);
\end{tikzpicture}
\end{center}
\end{example}

\begin{proposition}
For each $S = (Rules_x, Initials)$,
then there is an FSM $M$ such that $M$ represents $S$ for an initial state $I \in Initials$.
\end{proposition}

\begin{definition}
A string $\rho$ {\bf reflects} an execution $e = (s_1,a_1,p,a_2,s_2,t)$ 
iff $\rho$ is the string $\tau_1\ldots\tau_{t-1}$ and for each $1 \leq n < t$, $\tau_{n}$ is the tuple $(a_1(n),a_2(n))$.
\end{definition}

\begin{proposition}
Let $S = (Rules_x, Initials)$ be a system.
and let $M$ be an FSM that represents $S$ for $I \in Initials$.
\begin{enumerate}
\item for all $\rho$ s.t. $M$ accepts $\rho$, there is an $e$ s.t. $S$ generates $e$ and $e(0) = I$ and $\rho$ reflects $e$,
\item for all finite $e$ s.t. $S$ generates $e$ and $e(0) = I$, then there is a $\rho$ such that $M$ accepts $\rho$ and $\rho$ reflects $e$.
\end{enumerate}
\end{proposition}

So for each initial state for a system, we can obtain an FSM that is a concise representation of the executions of the system for that initial state. In Figure \ref{fig:algorithm}, we provide an  algorithm for generating these FSMs.
We show correctness for the algorithm as follows.

\begin{proposition}
Let $S$ = $(Rules_x,Initials)$ be a system
and let $I \in Initials$. 
If $M$ represents $S$ w.r.t. $I$ 
and ${\sf BuildMachine}(Rules_x,I)$ = $M'$, 
then $M = M'$.
\end{proposition}

An FSM provides a more efficient representation of all the possible executions than the set of executions for an initial state.
For instance, if there is a set of states that appear in some permutation of each of the executions then this can be more compactly represented by an FSM. And if there are infinite sequences, then again this can be more compactly represented by an FSM.

Once we have an FSM of a system with an initial state, we can ask obvious simple questions such as 
is termination possible, 
is termination guaranteed, and is one system subsumed by another?
So by translating a system into an FSM, we can harness substantial theory and tools for analysing FSMs.

Next we give a couple of very simple examples of FSMs obtained from executable logic. 
In these examples, we assume that agent 1 is trying to win an argument with agent 2. 
We assume that agent 1 has a goal. This is represented by the predicate ${\tt g(c)}$ in the private state of agent 1 for some argument ${\tt c}$. In its private state, each agent has zero or more arguments represented by the predicate ${\tt n(c)}$, and zero or more attacks ${\tt e(d,c)}$ from ${\tt d}$ to ${\tt c}$. In the public state, each argument $\tt c$ is represented by the predicate ${\tt a(c)}$.  Each agent can add attacks ${\tt e(d,c)}$ to the public state, if the attacked argument is already in the public state (i.e. ${\tt a(c)}$ is in the public state), and the agent also has the attacker in its private state (i.e. ${\tt n(d)}$ is in the private state). We have encoded the rules so that after an argument has been used as an attacker, it is removed from the private state of the agent so that it does not keep firing the action rule (this is one of a number of ways that we can avoid repetition of moves).

\begin{figure*}
\begin{center}
\begin{tikzpicture}
\node[state,initial] (s1) at (0,1) {$\sigma_1$};
\node[state] (s2) at (1.5,1) {$\sigma_2$};
\node[state] (s3) at (3,1) {$\sigma_3$};
\node[state] (s4) at (4.5,1) {$\sigma_4$};
\node[state] (s5) at (6,1) {$\sigma_5$};
\node[state,accepting] (s6) at (7.5,1) {$\sigma_6$};
\path[->] (s1) edge node[above] {$\tau_1$} (s2);
\path[->] (s2) edge node[above] {$\tau_2$} (s3);
\path[->] (s3) edge node[above] {$\tau_3$} (s4);
\path[->] (s4) edge node[above] {$\tau_4$} (s5);
\path[->] (s5) edge node[above] {$\tau_4$} (s6);
\end{tikzpicture}
\end{center}
\[
\begin{array}{c}
\footnotesize \sigma_1 = \tt (1, \{ g(a),n(a),n(c),e(c,b) \},\{  \}, \{n(b),e(b,a) \} ) \\
\footnotesize \sigma_2 = \tt (2, \{ g(a),n(c),e(c,b) \},\{ a(a) \}, \{n(b),e(b,a) \} )  \\
\footnotesize \sigma_3 = \tt (1, \{ g(a),n(c),e(c,b) \},\{ a(a),a(b,a) \}, \{ e(b,a) \} )   \\
\footnotesize \sigma_4 = \tt (2, \{ g(a),e(c,b) \},\{ a(a),a(b),a(c),a(c,b),a(b,a) \}, \{ e(b,a) \} )   \\
\footnotesize \sigma_5 = \tt (1, \{ g(a),e(c,b) \},\{ a(a),a(b),a(c),a(c,b),a(b,a) \}, \{ e(b,a) \} )  \\
\footnotesize \sigma_6 = \tt (0, \{ g(a),e(c,b) \},\{ a(a),a(b),a(c),a(c,b),a(b,a) \}, \{ e(b,a) \} )  \\
\\
\footnotesize \tau_1 = \tt (\{ \boxplus a(a), \ominus n(a) \}, \emptyset )\\ 
\footnotesize \tau_2 = \tt (\emptyset,\{\tt \boxplus a(b,a), \ominus n(b) \})\\
\footnotesize \tau_3 = \tt (\{ \boxplus a(c,b), \ominus n(c) \},\emptyset)\\
\footnotesize \tau_4 = \tt (\emptyset,\emptyset)\\
\end{array}
\]
\caption{\label{fig:fsm:1}The FSM for Example \ref{ex:fsm:1}}
\end{figure*}
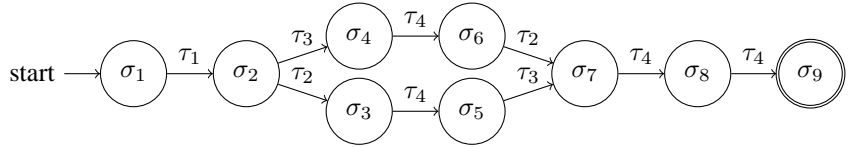

\begin{example}
\label{ex:fsm:1}
For the following action rules, with the initial state where the private state of agent 1 is $\{\tt g(a),n(a),n(c),e(c,b) \}$,
the public state is empty, and the private state of agent 2 is $\{\tt n(b),e(b,a) \})$, 
we get the FSM in Figure \ref{fig:fsm:1}. 
\[
\begin{array}{l}
\tt g(a) \wedge n(a) \Rightarrow \boxplus a(a) \wedge \ominus n(a)\\
\tt a(a) \wedge n(b) \wedge e(b,a) \Rightarrow \boxplus a(b,a) \wedge \ominus n(b)\\
\tt a(b) \wedge n(c) \wedge e(c,b) \Rightarrow \boxplus a(c,b) \wedge \ominus n(c)\\
\end{array}
\]
The terminal state therefore contains the following argument graph.
\begin{center}
\begin{tikzpicture}[->,>=latex,thick]
\node (a) [shape=rectangle,draw] at (4,1) {$a$};
\node (b) [shape=rectangle,draw] at (2,1) {$b$};
\node (c) [shape=rectangle,draw] at (0,1) {$c$};		
\path	(b) edge node[] {} (a);
\path	(c) edge node[] {} (b);
\end{tikzpicture}
\end{center}
Hence the goal argument $\tt a$ is in the grounded extension of the graph (as defined in \cite{Dun95}).
\end{example}

\begin{example}
\label{ex:fsm:2}
For the following action rules, with the initial state where the private state of agent 1 is $\{\tt g(a),n(a) \}$,
the public state is empty, and the private state of agent 2 is $\{\tt n(b),n(c),e(b,a),e(c,a) \})$, 
we get the FSM in Figure \ref{fig:fsm:2}
\[
\begin{array}{l}
\tt g(a) \wedge n(a) \Rightarrow \boxplus a(a) \wedge \ominus n(a)\\
\tt a(a) \wedge n(b) \wedge e(b,a) \Rightarrow \boxplus a(b,a) \wedge \ominus n(b)\\
\tt a(a) \wedge n(c) \wedge e(c,a) \Rightarrow \boxplus a(c,a) \wedge \ominus n(c)\\
\end{array}
\]
The terminal state therefore contains the following argument graph.
\begin{center}
\begin{tikzpicture}[->,>=latex,thick]
\node (b) [shape=rectangle,draw] at (4,1) {$b$};
\node (a) [shape=rectangle,draw] at (2,1) {$a$};
\node (c) [shape=rectangle,draw] at (0,1) {$c$};		
\path	(b) edge node[] {} (a);
\path	(c) edge node[] {} (a);
\end{tikzpicture}
\end{center}
Hence the goal argument $\tt a$ is in the grounded extension of the graph.
\end{example}

\begin{figure*}
\begin{center}
\begin{tikzpicture}
\node[state,initial] (s1) at (0,0.5) {$\sigma_1$};
\node[state] (s2) at (1.5,0.5) {$\sigma_2$};
\node[state] (s3) at (3,0) {$\sigma_3$};
\node[state] (s4) at (3,1) {$\sigma_4$};
\node[state] (s5) at (4.5,0) {$\sigma_5$};
\node[state] (s6) at (4.5,1) {$\sigma_6$};
\node[state] (s7) at (6,0.5) {$\sigma_7$};
\node[state] (s8) at (7.5,0.5) {$\sigma_8$};
\node[state,accepting] (s9) at (9,0.5) {$\sigma_9$};
\path[->] (s1) edge node[above] {$\tau_1$} (s2);
\path[->] (s2) edge node[above] {$\tau_2$} (s3);
\path[->] (s2) edge node[above] {$\tau_3$} (s4);
\path[->] (s3) edge node[above] {$\tau_4$} (s5);
\path[->] (s4) edge node[above] {$\tau_4$} (s6);
\path[->] (s5) edge node[above] {$\tau_3$} (s7);
\path[->] (s6) edge node[above] {$\tau_2$} (s7);
\path[->] (s7) edge node[above] {$\tau_4$} (s8);
\path[->] (s8) edge node[above] {$\tau_4$} (s9);
\end{tikzpicture}
\end{center}
\[
\begin{array}{c}
\footnotesize \sigma_1 = \tt (1, \{ g(a),n(a) \},\{  \}, \{ n(b),n(c),e(b,a),e(c,a) \} )   \\
\footnotesize \sigma_2 = \tt (2, \{ g(a) \},\{ a(a) \}, \{ n(b),n(c),e(b,a),e(c,a) \} ) \\
\footnotesize \sigma_3 = \tt (1, \{ g(a) \},\{ a(a),a(b),a(b,a) \}, \{ n(c),e(b,a),e(c,a) \} )  \\
\footnotesize \sigma_4 = \tt (1, \{ g(a) \},\{ a(a),a(c),a(c,a) \}, \{ n(b),e(b,a),e(c,a) \} )  \\
\footnotesize \sigma_5 = \tt (2, \{ g(a) \},\{ a(a),a(b),a(b,a) \}, \{ n(c),e(b,a),e(c,a) \} )   \\
\footnotesize \sigma_6 = \tt (2, \{ g(a) \},\{ a(a),a(c),a(c,a) \}, \{ n(b),e(b,a),e(c,a) \} ) \\
\footnotesize \sigma_7 = \tt (1,\{ g(a) \},\{ a(a),a(b),a(c),a(c,a),a(b,a) \}, \{ e(b,a),e(c,a)\} ) \\
\footnotesize \sigma_8 = \tt (2, \{ g(a) \},\{ a(a),a(b),a(c),a(c,a),a(b,a) \}, \{ e(b,a),e(c,a)\} ) \\
\footnotesize \sigma_9 = \tt (0, \{ g(a) \},\{ a(a),a(b),a(c),a(c,a),a(b,a) \}, \{ e(b,a),e(c,a) \} ) \\
\\
\footnotesize \tau_1 = (\tt \{ \boxplus a(a), \ominus n(a)\},\emptyset) )\\
\footnotesize \tau_2 = (\tt \emptyset, \{ \boxplus a(b,a), \ominus n(b) )\\
\footnotesize \tau_3 = (\tt \emptyset, \{ \boxplus a(c,a), \ominus n(c) )\\
\footnotesize \tau_4 = (\tt \emptyset,\emptyset)\\
\end{array}
\]
\caption{\label{fig:fsm:2}The FSM for Example \ref{ex:fsm:2}}
\end{figure*}


\begin{figure*}[]
\footnotesize
\[
\begin{array}{ll}
01 & {\sf BuildMachine}(Rules_x,I)\\
02 & \hspace{0.25cm} Start = (1,S_1,P,S_2) \mbox{ where } I = (S_1,A_1,P,A_2,S_2)\\
03 & \hspace{0.25cm} States_1 = NewStates_1 = \{Start\} \\
04 & \hspace{0.25cm} States_2 = Trans_1 = Trans_2 = \emptyset\\
05 & \hspace{0.25cm} x = 1, y = 2\\
06 & \hspace{0.25cm} {\sf While } \; NewStates_x \neq \emptyset\\ 
07 & \hspace{0.5cm} NextStates = NextTrans = \emptyset\\
08 & \hspace{0.5cm} {\sf For} \; (x,S_1,P,S_2) \in NewStates_x\\
09 & \hspace{1cm} Fired = \{ \psi \mid \phi\Rightarrow\psi \in Rules_x \mbox{ and } S_x\cup P \models \phi \} \\
10 &  \hspace{1cm} {\sf If } Fired == \emptyset\\
11 &  \hspace{1cm} {\sf Then } \; NextTrans = NextTrans \cup \{((x,S_1,P,S_2),(\emptyset,\emptyset),(y,S_1,P,S_2))\}\\
12 &  \hspace{1cm} {\sf Else \hspace{0.1cm} for } A \in {\sf Disjuncts}(Fired)\\
13 & \hspace{1.25cm} NewS = S_x \setminus \{ \alpha \mid \ominus\alpha \in A \} \cup \{ \alpha \mid \oplus\alpha\in A \}\\
14 & \hspace{1.25cm} NewP = P \setminus \{ \alpha \mid \boxminus\alpha \in  A \} \cup \{ \alpha \mid \boxplus\alpha\in A \}\\
15 &  \hspace{1.25cm} {\sf If } x == 1, NextState = (2,NewS,P,S_2) \mbox{ and } Label = (A,\emptyset) \\
16 &  \hspace{1.25cm} {\sf Else } \hspace{0.1cm} NextState = (1,S_1,P,NewS) \mbox{ and } Label = (\emptyset,A)  \\
17 & \hspace{1.25cm} NextStates = NextStates \cup \{ NextState \} \\
18 & \hspace{1.25cm} NextTrans = NextTrans \cup \{  ((x,S_1,P,S_2),Label,NextState) \} \\
19 & \hspace{0.5cm} {\sf If} \; x == 1, \mbox{\sf then } x = 2 \mbox{ and } y = 1, \mbox{\sf else } x = 1 \mbox{ and } y = 2\\
20 & \hspace{0.5cm} NewStates_x = NextStates\setminus States_x\\
21 & \hspace{0.5cm} States_x = States_x \cup NextStates\\
22 & \hspace{0.5cm} Trans_x = Trans_x \cup NextTrans\\
23 & \hspace{0.25cm} Close = \{  \sigma'' \mid (\sigma,\tau,\sigma'), (\sigma',\tau,\sigma'' ) \in Trans_1 \cup Trans_2  \}\\
24 & \hspace{0.25cm} Trans = {\sf MarkTrans}(Trans_1 \cup Trans_2, Close )\\
25 & \hspace{0.25cm} States = {\sf MarkStates}(States_1 \cup States_2, Close)\\
26 & \hspace{0.25cm} Term = {\sf MarkTerm}(Close)\\
27 & \hspace{0.25cm} Alphabet = \{ \tau \mid (\sigma,\tau,\sigma') \in States \}\\
28 & \hspace{0.25cm} {\sf Return } \; ( States, Trans, Start, Term, Alphabet  ) \\
\end{array}
\]
\caption{\label{fig:algorithm}
An algorithm for generating an FSM from a system $S$ = $(Rules_x,Initials)$ and an initial state $I$. 
The subsidiary function ${\sf Disjuncts}(Fired)$ is 
$\{ \{ \psi^1_1,..,\psi^1_{k_1} \}, ..,\{ \psi^i_1,..,\psi^1_{k_i} \}  
			\mid ((\psi^1_1\wedge..\wedge\psi^1_{k_1})\vee ..\vee (\psi^i_1\wedge..\wedge\psi^1_{k_i})) \in           Fired)   \}$.
For turn-taking, for agent $x$, $State_x$ is the set of expanded states and $NewStates_x$ is the set of unexpanded states.
Lines 02-05 set up the construction with agent 1 being the agent to expand the initial state.
At lines 06-18, when it is turn of $x$, each unexpanded state in $NewStates_x$ is expanded by identifying the fired rules.
At lines 10-11, if there are no fired rules, then the empty transition (i.e. $(\emptyset,\emptyset)$) is obtained,
otherwise at lines 12-17, each disjunct for each fired rule gives a next state and transition that is added to $NextStates$ and $NextTrans$ accordingly.
At lines 19-22, the turn is passed to the other agent, 
and $NewStates_x$, $States_x$, and $Trans_x$ updated.
At line 23, the terminal states are identified from the transitions.
At line 24, the ${\sf MarkTrans}$ function returns the union of the transitions for each agent but for each $\sigma = (x,S1,P,S2) \in Term$, $\sigma$ is changed to $(0,S1,P,S2)$ in order to mark it as a terminal state in the FSM.
At line 25, the ${\sf MarkStates}$ function returns the union of the states for each agent but for each $\sigma = (x,S1,P,S2) \in Term$, $\sigma$ is changed to $(0,S1,P,S2)$, 
and similarly at line 26, ${\sf MarkTerm}$ function returns the set $Close$ but with each state being of the form $(0,S1,P,S2)$.
}
\end{figure*}

In the above examples, we have considered a formalisation of dialogical  argumentation where agents exchange abstract arguments and attacks. It is straightforward to formalize other kinds of example to exchange a wider range of moves, richer content (e.g. logical arguments composed of premises and conclusion \cite{parsons:03a}), and richer notions (e.g. value-based argumentation \cite{Ben03}).


\section{Minimax analysis of finite state machines}\label{section:analysis}

Minimax analysis is applied to two-person games for deciding which moves to make. 
We assume two players called MIN and MAX. 
MAX moves first, and they take turns until the game is over. 
An {\bf end function} determines when the game is over. Each state where the game has ended is an {\bf end state}.
A {\bf utility function} (i.e. a payoff function) gives the outcome of the game (eg chess has win, draw, and loose).
The {\bf minimax strategy} is that MAX aims to get to an end state that maximizes its utility regardless of what MIN does

We can apply the minimax strategy to the FSM machines generated for dialogical argumentation as follows:
(1) Undertake breadth-first search of the FSM; 
(2) Stop searching at a node on a branch if the node is an end state according to the end function (note, this is not necessarily a terminal state in the FSM); 
(3) Apply the utility function to each leaf node $n$ (i.e. to each end state) in the search tree to give the value $value(n)$ of the node; 
(4) Traverse the tree in post-order, and calculate the value of each non-leaf node as follows where the non-leaf node $n$ is at depth $d$ and with children $\{ n_1,..,n_k\}$:
\begin{itemize}
\item If $d$ is odd, then $value(n)$ is the maximum of $value(n_1)$,.., $value(n_k)$.
\item If $d$ is even, then $value(n)$ is the minimum of $value(n_1)$,.., $value(n_k)$.
\end{itemize}

There are numerous types of dialogical argumentation that can be modelled using propositional executable logic and analysed using the minimax strategy. Before we discuss some of these options, we consider some simple examples where we assume that the search tree is exhaustive, (so each branch only terminates when it reaches a terminal state in the FSM), and the utility function returns 1 if the goal argument is in the grounded extension of the graph in the terminal state, and returns 0 otherwise.

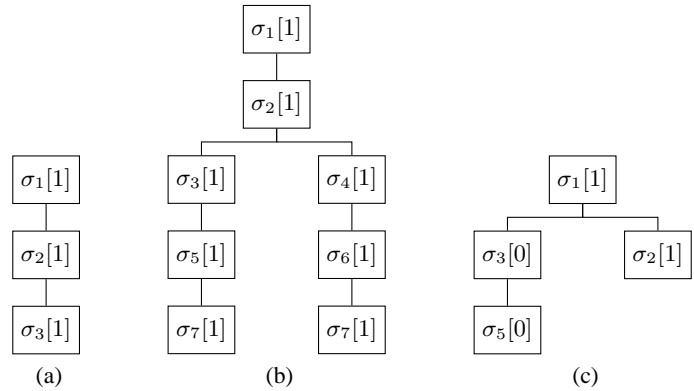
\begin{figure*}[]
\begin{center}
\subfloat[\label{fig:minimax:a}] {\footnotesize
\begin{tikzpicture}[every node/.style={draw,rectangle,minimum height=2em,minimum width=2em},
level distance=10mm,
level 1/.style={sibling distance=40mm},
level 2/.style={sibling distance=20mm},
level 3/.style={sibling distance=10mm}]
\node { $ \sigma_1 [1] $  }
[edge from parent fork down]
	child {node {$ \sigma_2  [1]  $}
			child {node {$ \sigma_3 [1]   $}}
			};
\end{tikzpicture}
}
\hspace{1cm}
\subfloat[\label{fig:minimax:b}] {\footnotesize
\begin{tikzpicture}[every node/.style={draw,rectangle,minimum height=2em,minimum width=2em},
level distance=10mm,
level 1/.style={sibling distance=20mm},
level 2/.style={sibling distance=20mm},
level 3/.style={sibling distance=10mm}]
\node { $ \sigma_1 [1] $  }
[edge from parent fork down]
	child {node {$ \sigma_2   [1] $}
			child {node {$ \sigma_3  [1] $}
					child {node {$ \sigma_5  [1] $}
							child {node {$ \sigma_7 [1]  $}}				
							}
					} 
			child {node {$ \sigma_4  [1]$}
					child {node {$ \sigma_6 [1]  $}
							child {node {$ \sigma_7 [1]  $}}
							}
					}
			};
\end{tikzpicture}
}
\hspace{1cm}
\subfloat[\label{fig:minimax:c}] {\footnotesize
\begin{tikzpicture}[every node/.style={draw,rectangle,minimum height=2em,minimum width=2em},
level distance=10mm,
level 1/.style={sibling distance=20mm},
level 2/.style={sibling distance=20mm},
level 3/.style={sibling distance=10mm}]
\node { $ \sigma_1 [1] $  }
[edge from parent fork down]
	child {node {$ \sigma_3   [0] $}
			child {node {$ \sigma_5 [0]  $}}
			}			
	child {node {$ \sigma_2 [1]   $}
			};
\end{tikzpicture}
}
\end{center}
\caption{\label{fig:minimaxtree}
Minimax trees for Examples \ref{ex:tree:1} and \ref{ex:fsm:3}. 
Since each terminal state in an FSM is a copy of the previous two states, we save space by not giving these copies in the search tree.
The minimax value for a node is given in the square brackets within the node. (a) is for Example \ref{ex:fsm:1}, (b) is for Example \ref{ex:fsm:2} and (c) is for Example \ref{ex:fsm:3}}
\end{figure*}

\begin{example}
\label{ex:tree:1}
From the FSM in Example \ref{ex:fsm:1}, we get the minimax search tree in Figure \ref{fig:minimax:a},
and from the FSM in Example \ref{ex:fsm:2}, we get the minimax search tree in Figure \ref{fig:minimax:b}.
In each case, the terminal states contains an argument graph in which the goal argument is in the grounded extension of the graph.
So each leaf of the minimax tree has a utility of 1, and each non-node has the value 1. 
Hence, agent 1 is guaranteed to win each dialogue whatever agent 2 does.
\end{example}

The next example is more interesting from the point of view of using the minimax strategy since agent 1 has a choice of what moves it can make and this can affect whether or not it wins.

\begin{example}
\label{ex:fsm:3}
In this example, we assume agent 1 has two goals $\tt a$ and $\tt b$, but it can only present arguments for one of them. So if it makes the wrong choice it can loose the game.
The executable logic rules are given below and the resulting FSM is given in Figure \ref{fig:fsm:3}.
For the minimax tree (given in Figure \ref{fig:minimax:c}) the left branch results in an argument graph in which the goal is not in the grounded extension, whereas the right branch terminates in an argument graph in which the goal is in the grounded extension. By a minimax analysis, agent 1 wins. 
\[
\begin{array}{l}
\tt g(a) \wedge n(a) \Rightarrow \boxplus a(a) \wedge \ominus n(a) \wedge \ominus g(b) \\
\tt g(b) \wedge n(b) \Rightarrow \boxplus a(b) \wedge \ominus n(b) \wedge \ominus g(a) \\
\tt a(a) \wedge n(c) \wedge e(c,a) \Rightarrow \boxplus a(c,a) \wedge \ominus n(c)\\
\end{array}
\]
\end{example}

\begin{figure*}
\begin{center}
\begin{tikzpicture}
\node[state,initial] (s1) at (0,0.5) {$\sigma_1$};
\node[state] (s2) at (1.5,1) {$\sigma_2$};
\node[state] (s4) at (3,1) {$\sigma_4$};
\node[state,accepting] (s6) at (4.5,1) {$\sigma_6$};
\node[state] (s3) at (1.5,0) {$\sigma_3$};
\node[state] (s5) at (3,0) {$\sigma_5$};
\node[state] (s7) at (4.5,0) {$\sigma_7$};
\node[state,accepting] (s8) at (6,0) {$\sigma_8$};

\path[->] (s1) edge node[above] {$\tau_1$} (s2);
\path[->] (s1) edge node[above] {$\tau_2$} (s3);

\path[->] (s2) edge node[above] {$\tau_4$} (s4);

\path[->] (s3) edge node[above] {$\tau_3$} (s5);
\path[->] (s4) edge node[above] {$\tau_4$} (s6);
\path[->] (s5) edge node[above] {$\tau_4$} (s7);

\path[->] (s7) edge node[above] {$\tau_4$} (s8);
\end{tikzpicture}
\end{center}
\[
\begin{array}{c}
\footnotesize \sigma_1 = \tt (1, \{ g(a),g(b),n(a),n(b) \},\{  \}, \{ n(c),e(c,a) \} )   \\
\footnotesize \sigma_2 = \tt (2, \{ g(a),g(b),n(a) \},\{ a(b) \}, \{ n(c),e(c,a) \} )   \\
\footnotesize \sigma_3 = \tt (2, \{ g(a),g(b),n(b) \},\{ a(a) \}, \{ n(c),e(c,a) \} )  \\
\footnotesize \sigma_4 = \tt (1, \{ g(a),g(b),n(a) \},\{ a(b) \}, \{ n(c),e(c,a) \} )   \\
\footnotesize \sigma_5 = \tt (1, \{ g(a),g(b),n(b) \},\{ a(a),a(c),a(c,a) \}, \{ e(c,a)\} )  \\
\footnotesize \sigma_6 = \tt (0, \{ g(a),g(b),n(a) \},\{ a(b) \}, \{ n(c),e(c,a) \} ) \\
\footnotesize \sigma_7 = \tt (2, \{ g(a),g(b),n(b) \},\{ a(a),a(c),a(c,a) \}, \{ e(c,a)\} )  \\
\footnotesize \sigma_8 = \tt (0, \{ g(a),g(b),n(b) \},\{ a(a),a(c),a(c,a) \}, \{ e(c,a)\} )  \\
\\
\footnotesize \tau_1 = ( \{ \tt  \boxplus a(b), \ominus n(b), \ominus g(a)   \}, \emptyset )\\
\footnotesize \tau_2=  ( \{ \tt \boxplus a(a), \ominus n(a), \ominus g(b) \}, \emptyset)\\
\footnotesize \tau_3 = ( \emptyset, \{ \tt  \boxplus a(c,a), \ominus n(c)   \} )\\
\footnotesize \tau_4 =  ( \emptyset, \emptyset )\\
\end{array}
\]
\caption{\label{fig:fsm:3}The FSM for Example \ref{ex:fsm:3}}
\end{figure*}


We can use any criterion for identifying the end state. In the above, we have used the {\bf exhaustive end function} giving an end state (i.e. the leaf node in the search tree) which is a  terminal state in the FSM followed by two empty transitions. If the branch does not come to a terminal state in the FSM, then it is an infinite branch.  We could use a {\bf non-repetitive end function} where the search tree stops when there are no new nodes to visit. 
For instance, for example \ref{ex:fsmcycle}, we could use the non-repetitive end function to give a search tree that contains one branch $\sigma_1,\sigma_2,\sigma_3$ where $\sigma_1$ is the root and $\sigma_3$ is the leaf. 
Another simple option is a {\bf fixed-depth end function} which has a specified maximum depth for any branch of the search tree. 
More advanced options for end functions include {\bf concession end function} when an agent has a loosing position, and it knows that it cannot add anything to change the position, then it concedes.

There is also a range of options for the utility function. In the examples, we have used grounded semantics to determine whether a goal argument is in the grounded extension of the argument graph specified in the terminal public state.  A refinement is the {\bf weighted utility function} which weights the utility assigned by the grounded utility function by $1/d$ where $d$ is the depth of the leaf.  The aim of this is to favour shorter dialogues. Further definitions for utility functions arise from using other semantics such as preferred or stable semantics and richer formalisms such as valued-based argumentation \cite{Ben03}.

\section{Implementation study}\label{section:implementation}

In this study, we have implemented three algorithms: The generator algorithm for taking an initial state and a set of action rules for each agent, and outputting the fabricated FSM; A breadth-first search algorithm for taking an FSM and a choice of termination function, and outputting a search tree; And a minimax assignment algorithm for taking a search tree and a choice of utility function, and outputting a minimax tree. These implemented algorithms were used together so that given an initial state and rules for each agent, the overall output was a minimax tree. This could then be used to determine whether or not agent 1 had a winning strategy (given the initial state). The implementation incorporates the exhaustive termination function, and two choices of utility function (grounded and weighted grounded). 

The implementation is in Python 2.6 and was run on a Windows XP PC with Intel Core 2 Duo CPU E8500 at 3.16 GHz and 3.25 GB RAM. 
For the evaluation, we also implemented an algorithm for generating tests inputs. Each test input comprised an initial state, and a set of action rules for each agent. Each initial state involved 20 arguments randomly assigned to the two agents and up to 20 attacks per agent. For each attack in an agent's private state, the attacker is an argument in the agent's private state, and the attacked argument is an argument in the other agent's private state. The results are presented in Table \ref{tab:results}. 

\begin{table*}
\begin{center}
{\small
\begin{tabular}{| r| r| r| r| r| r| r|}
\hline
Average no.& Average no.& Average no. & Average no.&  Average& Median & No. of runs\\
attacks & FSM nodes & FSM transitions & tree nodes & run time & run time & timed out\\
\hline
9.64   & 6.29& 9.59   & 31.43 &  0.27 & 0.18 & 0 \\
11.47 & 16.01& 39.48  & 1049.14 &  6.75 &  0.18&  1\\
13.29 & 12.03 & 27.74 &  973.84 &  9.09 &  0.18&  2\\
14.96 &  12.50& 27.77 & 668.65 &  6.41 &  0.19&  13\\
16.98 &  19.81& 49.96 &  2229.64 &  25.09 &  0.20 &  19\\
18.02 &   19.01 & 47.81  &  2992.24 &  43.43 &  0.23 &  30\\
\hline
\end{tabular}}
\end{center}
\caption{\label{tab:results}The results from the implementation study. Each row is produced from 100 runs. 
Each run (i.e. a single initial state and action rules for each agent) was timed. If the time exceeded 100 seconds for the generator algorithm, the run was terminated}
\end{table*}

As can be seen from these results, up to about 15 attacks per agent, the implementation runs in negligible time. However, above 15 attacks per agent, the time did increase markedly, and a substantially minority of these timed out. To indicate the size of the larger FSMs, consider the last line of the table where the runs had an average of 18.02 attacks per agent: For this set, 8 out of 100 runs had 80+ nodes in the FSM. Of these 8 runs, the number of states was between 80 and 163, and the number of transitions was between 223 and 514.

The algorithm is somewhat naive in a number of respects. For instance, the algorithm for finding the grounded extension considers every subset of the set of arguments (i.e. $2^{20}$ sets). Clearly more efficient algorithms can be developed or calculation subcontracted to a system such as ASPARTIX \cite{Aspartix08}. Nonetheless, there are interesting applications where 20 arguments would be a reasonable, and so we have shown that we can analyse such situations successfully using the Minimax strategy, and with some refinement of the algorithms, it is likely that larger FSMs can be constructed and analysed. 

Since the main aim was to show that FSMs can be generated and analysed, we only used a simple kind of argumentation dialogue.
It is straightforward to develop alternative and more complex scenarios, using the language of propositional executable logic e.g. for capturing beliefs, goals, uncertainty etc, for specifying richer behaviour.

\section{Discussion}\label{section:discussion}

In this paper, we have investigated a uniform way of presenting and executing dialogical argumentation systems based on a propositional executable logic. As a result different dialogical argumentation systems can be compared and implemented more easily than before. The implementation is generic in that any action rules and initial states can be used to generate the FSM and properties of them can be identified empirically. 

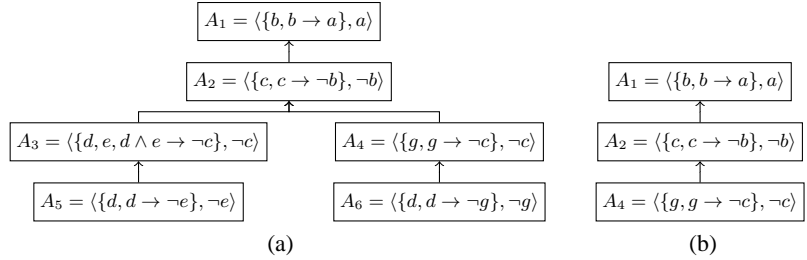
\begin{figure*}
\begin{center}
\subfloat[\label{fig:logic:left}] {\footnotesize
\begin{tikzpicture}[every node/.style={draw,rectangle,minimum height=2em,minimum width=2em,scale=0.8},
level distance=8mm,
level 1/.style={sibling distance=10mm},
level 2/.style={sibling distance=40mm},
level 3/.style={sibling distance=10mm}]
\node[] { $A_1 = \langle \{ b, b \rightarrow a \}, a \rangle  $  }
[edge from parent fork down,<-]
	child {node[] {$A_2 =   \langle \{ c, c \rightarrow \neg b \}, \neg b \rangle  $}
			child {node[] {$A_3 =  \langle \{ d,e, d \wedge e \rightarrow \neg c \}, \neg c \rangle  $}
					child {node[] {$A_5 =  \langle \{ d, d \rightarrow \neg e \}, \neg e \rangle  $}}
					}
			child {node[] {$A_4 =  \langle \{ g, g \rightarrow \neg c \}, \neg c \rangle  $}
					child {node[] {$A_6 =  \langle \{ d, d \rightarrow \neg g \}, \neg g \rangle  $}}
					}
			};
\end{tikzpicture}
}
\hspace{0.5cm}
\subfloat[\label{fig:logic:right}] {\footnotesize
\begin{tikzpicture}[every node/.style={draw,rectangle,minimum height=2em,minimum width=2em,scale=0.8},
level distance=8mm,
level 1/.style={sibling distance=40mm},
level 2/.style={sibling distance=50mm},
level 3/.style={sibling distance=10mm}]
\node[] { $A_1 = \langle \{ b, b \rightarrow a \}, a \rangle  $  }
[edge from parent fork down,<-]
	child {node[] {$A_2 =   \langle \{ c, c \rightarrow \neg b \}, \neg b \rangle  $}
			child {node[] {$A_4 =  \langle \{ g, g \rightarrow \neg c \}, \neg c \rangle  $}}
			};
\end{tikzpicture}
}
\end{center}
\caption{
Consider the following knowledgebases for each agent
$\Delta_1 = \{ b, d, e, g  , b \rightarrow a, d \wedge e \rightarrow \neg c, g \rightarrow \neg c   \}$
and $\Delta_2 = \{  c, c \rightarrow \neg b, d \rightarrow \neg e,  d \rightarrow \neg g    \}$.
(a) Agent 1 is exhaustive in the arguments posited, thereby allowing agent 2 to construct arguments that cause the root to be defeated.
(b)Agent is selective in the arguments posited, thereby ensuring that the root is undefeated.}
\end{figure*}

In the examples in this paper, we have assumed that when an agent presents an argument, the only reaction the other agent can have is to present a counterargument (if it has one) from a set that is fixed in advance of the dialogue. Yet when agents argue, one agent can reveal information that can be used by the other agent to create new arguments. We illustrate this in the context of logical arguments. Here, we assume that each argument is a tuple $\langle\Phi,\psi\rangle$ where $\Phi$ is a set of formulae that entails a formula $\psi$. In Figure \ref{fig:logic:left}, we see an argument graph instantiated with logical arguments. Suppose arguments $A_1$, $A_3$ and $A_4$ are presented by agent 1, and arguments $A_2$, $A_5$ and $A_6$ are presented by agent 2. Since agent 1 is being exhaustive in the arguments it presents, agent 2 can get a formula that it can use to create a counterargument. In Figure \ref{fig:logic:right}, agent 1 is selective in the arguments it presents, and as a result, agent 2 lacks a formula in order to construct the counterarguments it needs. We can model this argumentation in propositional executable logic, generate the corresponding FSM, and provide an analysis in terms of minimax strategy that would ensure that agent 1 would provide $A_4$ and not $A_3$, thereby ensuring that it behaves more intelligently. We can capture each of these arguments as a proposition and use the minimax strategy in our implementation to obtain the tree in Figure \ref{fig:logic:right}. 

General frameworks for dialogue games have been proposed \cite{ME98,MP02b}. They offer insights on dialogical argumentation systems, but they do not provide sufficient detail to formally analyse or implement specific systems. A more detailed framework, that is based on situation calculus, has been proposed by Brewka \cite{Bre01}, though the emphasis is on modelling the protocols for the moves made in dialogical argumentation based on the public state rather than on strategies based on the private states of the agents.


The minimax strategy has been considered elsewhere in models of argumentation (such as for determining argument strength \cite{MT08} and for marking strategies for dialectical trees \cite{RMS09}, for deciding on utterances in a specific dialogical argumentation \cite{ON09}). However, this paper appears to be the first empirical study of using the minimax strategy in dialogical argumentation. 

In future work, we will extend the analytical techniques for imperfect games where only a partial search tree is constructed before the utility function is applied, and extend the representation with weights on transitions (e.g. weights based on tropical semirings to capture probabilistic transitions) to explore the choices of transition based on preference or uncertainty.



\end{document}